# Tuning-less Object Naming with a Foundation Model.

Andrej Lucny
*Department of Applied Informatics*
*Comenius University*
Bratislava, Slovakia
lucny@fmph.uniba.sk

Pavel Petrovic
*Department of Applied Informatics*
*Comenius University*
Bratislava, Slovakia
pavel.petrovic@fmph.uniba.sk

*Abstract*— We implement a real-time object naming system that enables learning a set of named entities never seen. Our approach employs an existing foundation model that we consider ready to see anything before starting. It turns seen images into relatively small feature vectors that we associate with index to a gradually built vocabulary without any training of fine-tuning of the model. Our contribution is using the association mechanism known from transformers as attention. It has features that support generalization from irrelevant information for distinguishing the entities and potentially enable associating with much more than indices to vocabulary. As a result, the system can work in a one-shot manner and correctly name objects named in different contents. We also outline implementation details of the system modules integrated by a blackboard architecture. Finally, we investigate the system's quality, mainly how many objects it can handle in this way.

## I. Introduction

How speakers refer to objects is a central issue of natural language processing and computer vision conjunction [1]. The traditional approach solves it by training models with parameters dedicated to tuning after each new object and its name [2] [3] [4]. Researchers seeking alternative solutions to various tasks that avoid this necessity employ foundation models [5] trained from vast unannotated data. We introduce an approach that uses one such general-purpose foundation model to implement a real-time system for object naming. Our contribution is that the system avoids any tuning and works in a one-shot manner: naming one instance of an object enables the system to name most of the object instances correctly.

The system we implement runs in real-time and works as follows: The user sits in front of the robot's camera and shows objects like a cup, knife, bottle, pen, watch, and others to it and names them like: "This is a cup." Robot (in our case, simulated iCub [6]) responds: "O.K." From that moment, the user can show any one of the named objects (in any position) and ask: "What is this?" And the robot answers, e.g., "This is a cup." If the robot fails to name the object correctly, we can fix its behavior by saying: "No, this is a bottle." Optionally we solicit that if we present an unknown entity, the robot responds: "I have no idea." (Figure 1)

The system stems from several pieces of available stuff. The following chapter describes what is necessary for understanding its operation. We mention the rest within the implementation details later. Finally, we present the results. Thanks to the capacities of the foundation model and the carefully selected task, the user finds a remarkable quality.

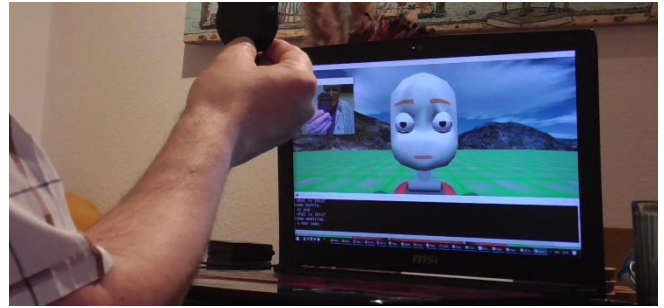

Fig. 1. Circumstances of our system for naming objects operation

## II. Related Works

We can track back the history of object naming via robots to the early days of robotics research. Traditionally, algorithms resided in handcrafted features and employed edge detection, texture analysis, and template matching techniques. These methods were limited in their ability to handle variations in object appearance. In recent years, deep learning has revolutionized object recognition in robotics and enabled incremental learning. Here we describe some key aspects and methods that contribute to our approach.

### A. Foundation model

We employ a pre-trained model called DINO ViT backbone (medium size) developed by Microsoft Research [7]. This model aims to see anything, as Chat GPT can answer any question. Therefore, we could consider it a large image model by analogy or - despite having no prompts - an early foundation model. Its architecture stems from the Visual transformer (ViT) [8] and turns images into rather-small feature vectors (mere 384 features for the medium size) (Figure 2).

Its training employed the DINO method. This method resides in contrastive learning with the metric based on the similarity of outputs of two copies of the same network fed with an image and its augmentation. We train only one copy (student) that we sometimes copy into the fixed copy (teacher). In this way, we gradually get a better and better feature extractor. Its quality is impressive. It maps images into a latent space of dimension 384, where each point corresponds to a potentially valid input from a camera. It maps similar images into close points; thus, a video corresponds to a likely trajectory. Since the primary purpose of the backbone model is to provide features for a bigger model trained (from annotated data) to provide attention maps, these features are related to the faculties of the most salient object in front of the camera.

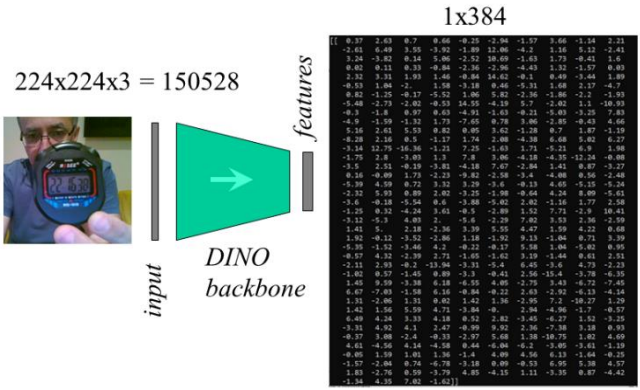

Fig. 2. Example of DINO ViT backbone model (medium size) I/O

When we move with an object in front of the camera, some features remain almost the same while others vary. This fact led us to the idea that for a number of objects significantly smaller than the number of features, it is possible to distinguish the features essential for their mutual distinction from those insignificant.

*B. Association mechanism*

The associating mechanism we use, known from transformers [9] (under the – not entirely suitable for us – name attention), works with a set of $l$ key-value pairs. When we have a query $q$ at the input, we try to mix the query from the keys $K$ and create the output as an analog mixture of the corresponding values $V$, where:

$$K = \begin{pmatrix} k_1 \\ k_2 \\ \dots \\ k_l \end{pmatrix} \quad V = \begin{pmatrix} v_1 \\ v_2 \\ \dots \\ v_l \end{pmatrix}$$

All queries and keys are vectors of dimension $n$, so $K$ is an $l \times n$ matrix. Values and outputs are vectors of dimension $m$, so $V$ is an $l \times m$ matrix. First we find $c_i \in <0,1>$ such that $\Sigma c_i = 1$, $\Sigma c_i k_i = pr_K(q)$, and $i=1,2,...l$, where $pr_K(q)$ is a vector similar to the projection of $q$ into the subspace generated by the keys $K$. We want $c_i$ to express the similarity between the key $k_i$ and the query $q$, t so we can derive it from the scalar product $q\, k_i$, proportional to the cosine of the angle that $q$ and $k_i$ make. However, we have to get these similarities - positive for congruent, zero for mutually perpendicular, and negative for opposite vectors - to the interval $<0,1>$, which can be arranged by the function *softmax* defines as $softmax(x)_i = exp(x_i) / \Sigma_k exp(x_k)$. The chosen coefficients with which we mix the keys $k_i$ into something similar to the query $q$ are, therefore:

$$c = softmax\left(\frac{qK^T}{d}\right)$$

where d is the scaling factor by which we determine how much we mix from similar keys and how much from different ones. The smaller this constant is, the closer the coefficients are to the so-called one-hot encoding (one unit and other zeros). For $d = 1/n$, where $n$ is the dimension of the keys, we practically always lean towards the dominance of one key, while the favorite value $d = \sqrt{n}$ ensures that we constantly mix a little from the other keys as well. A higher $d$ can benefit the association mechanism's ability to find a proper response even for queries for which no similar key we memorized, but we can consider them as transitional forms between two or more memorized keys. Now that we have the coefficients $c$, which roughly expresses the query as a mixture of the keys, we can analogically mix the values of $V$ to output $o = cV$. So the complete response of association mechanism $A$ to query $q$ is:

$$A(q, K, V) = softmax\left(\frac{qK^T}{d}\right)V$$

When the number of keys is much less than their dimension, the query can be expressed as a linear combination of keys only if it is lying in the hyperplane generated by the keys. We design the mechanism to provide the same response as for the linear projection of the query onto the hyperplane, in this case. (It is the fundamental feature of the scalar product.) It allows the generalization from irrelevant background conditions. Therefore, collecting the keys under the same background conditions is helpful, although we employ them later in a different situation.

*C. Integration via Blackboard*

Although we can call the used foundation model relatively quickly (every 50 ms on GPU), we cannot process every frame grabbed from the camera (30fps). Such a system must combine fast data sources like a camera, slower models, and languid robot movement. Therefore, the integration employs a blackboard architecture. Concretely, we use our solution named Agent-Space architecture [10] that split the system into a set of agents communicating via blackboard and let the overall control to be emerging from the individual behaviors of the agents.

```
from agentspace import space

space['a'] = 2
print(space['a']) # 2

print(space['b']) # None

print(space(default=-1)['b']) # -1

space(validity=0.5)['b'] = 1
print(space['b']) # 1
time.sleep(1)
print(space(default=-1)['b']) # -1

space['c'] = 1 # default priority is 1.0
space(validity=0.5,priority=2.0)['c'] = 2
space(validity=2.0,priority=1.5)['c'] = 3
print(space(default=0)['c']) # 2
time.sleep(1)
print(space(default=0)['c']) # 0
```

Fig. 3. Examples of calling backboard read/write operations

The agents communicate only indirectly through a backboard called space. Space is an extended dictionary containing variables (name-value) that enables one to define:

- default values for reading unavailable data
- time validity of the written data (after expiration, the data automatically disappear)
- priority of the writing operation (a lower-priority operation cannot overwrite valid data written by a higher-priority one) (Figure 3)

Within this architecture, we define agents by overriding two methods:

- initialization method called at the startup
- the method called periodically (by a timer) or upon the availability of new data (by a trigger) (we create timers or triggers during the initialization)

We run the agents in parallel; hence they are implemented as objects owning their thread. We design the system to start agents in any order.

Thanks to the features of this architecture, we can quickly implement quite complicated relationships among individual system modules.

## III. METHOD

The core of our method resides in associating image features [11] with the code of the index of a word in a vocabulary we gradually extend. As mentioned above, image features provided by the foundation model are suitable to be associated with other data. During the system operation, the user names a shown object. At that moment, we collect the key-value pair, where the key is the image feature vector, and the value represents the name's index (Figure 4). In this way, we can gradually name more and more concrete points of the image features latent space. When the user asks the robot to name a shown thing, the robot employs the association mechanism: It mixes the current image feature vector from those associated ones and calculates an analogical mixture from the indices representation. Then the robot interprets the result as an index into the vocabulary and names the shown object accordingly.

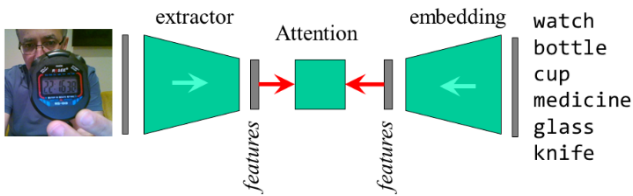

Fig. 4. Association mechanism

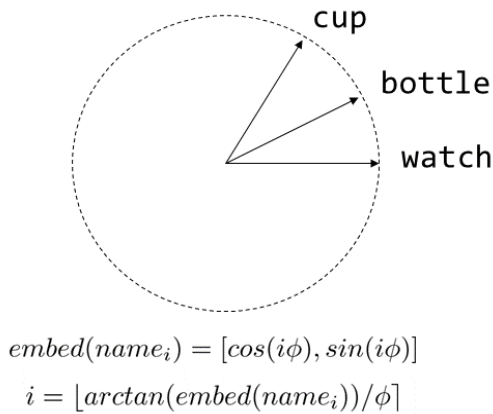

$$embed(name_i) = [cos(i\phi), sin(i\phi)]$$
$$i = \lfloor arctan(embed(name_i))/\phi \rceil$$

Fig. 5. Positional encoding of the vocabulary

Of course, we must adequately represent the indices pointing to the vocabulary. Coding them by integers is unsuitable since all value vectors would lie on the same line, and mixing values would always lead to a wrong index. On the other hand, if we avoid mixing and responding, e.g., with the value associated with the closest key, we would lose the ability to abstract from the irrelevant features, going outside the hyperplane generated by keys. Therefore, we must represent values by vectors of dimension at least two. The simplest solution is to put each word onto a unit circle with a given angle step $\varphi$ (we have used $\varphi = 0.2\ rad$). So, we represent index $i$ by vector $(cos\ i\varphi, sin\ i\varphi)$. It is nothing else than the simplest case of the so-called positional encoding (Figure 5).

Communication between the user and the robot employs natural speech. The robot listens and transcribes the recorded sound into a text by the Whisper model. Then, it analyzes text and recognizes commands corresponding to the information about the object name or the request to name it.

In the first case, we check if the name is already in the vocabulary. If yes, we find its index or extend the vocabulary and take the index of the last word otherwise. Then we associate the current image feature vector with the positional encoding of the index by adding such a pair to the key-values list. Finally, we call speech synthesis to let the robot respond "O.K."

In the second case, we apply the association mechanism to approximately mix the current image features from the known keys and calculate an analogical mixture from values. It provides us with a representation $(x,y)$ of an index that we can reveal as $i = atan(y/x)/\varphi$. We round the result to the closest integer. If it is not a valid index, we call speech synthesis to let the robot say: "I have no idea..." while the robot names the seen object otherwise.

For a better effect, we can add some movements of the robot, e.g., moving lips during speaking.

So, when we start the system and ask the robot to name an object, it responds that it has no idea. Then we call the first object, and the robot responds O.K. As a result, if we ask to name that object, the robot reacts correctly. However, it gives the same name to any presented entity. For proper response, we have to show things we have already called. Occasionally, it can happen that, e.g., rotation by 90 degrees causes the naming fails. In such a situation, we can say: "It is a ..." and supply the correct name that fixes the robot's reaction immediately and definitely. This way, we could collect more than one association pair for an object.

As an option, we could aim to let the robot recognize that we have not called an object yet. When we mix a query from keys, the coefficients of the mixture provided by the softmax function give no cue whether we are far or close to keys. However, the cosine similarities before applying softmax give us better information. We can compare their maximum. The higher it is, the closer we are to a particular key. Experimentally, we can estimate a threshold typical for the feature extractor and used objects. If the maximum is below the threshold, we let the robot say, "I have no idea," regardless, we have a valid index. Of course, avoiding the wrong names, we sometimes face "no idea" when presenting an already-named object. As a result, we need to call an object more times, and we have more key-value pairs for its name.

Finally, we still have a reserve in involving continuity of watching objects to collect additional association pairs automatically whenever the named entity recognition fails. In this case, we must be able to detect that the thing in front of the camera has changed. We can manage it by using the final product of the DINO model suite that deals with attention map detection, not just the backbone.

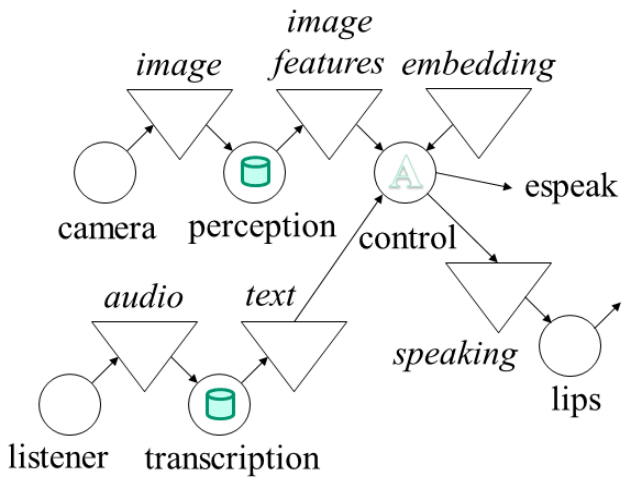

Fig. 6. The overall layout of our object naming system

IV. IMPLEMENTATION

Implementing our method, we had to respect that the corresponding real-time system must run several coordinated processes in parallel. Therefore, we employ the above blackboard architecture, enabling us to modularize the system comfortably. It contains the following modules (called agents) (Figure 6):

- *ListenerAgent* listens to the microphone and records the sound. When it detects silence, it outputs the recorded sound to the blackboard variable *audio* and listens again.
- *TranscriptionAgent* loads the Whisper model and registers the trigger on the variable *audio* during its initialization. Then wakes up when the sound recorded by the *ListenerAgent* is available and calls the Whisper model to transcribe them into text. Finally, it writes the text to the variable *text* on the blackboard and sleeps.
- *CameraAgent* grabs images from the camera and puts them into the variable *camera* on the blackboard.
- *PerceptionAgent* register trigger on the variable *camera*. Then it reads the image from the variable, feeds it into the foundation model, and outputs the obtained features to the variable *features*. Since the pictures provided by the camera flow through the blackboard, this agent does not process all of them. They are automatically sampled according to the model's ability to process them. The longer the model turns the image into the features, the less frequently it reads a new image.
- *ControlAgent* registers trigger on the variable *text*. When the TranscriptionAgent provides the text, this agent reads and analyzes it. If it recognizes a piece of naming information, it checks the presence of the name in the vocabulary. If it is a new name, the agent adds it to the list of words. Anyway turns its index into a vector by the positional encoding representation. Then reads the image feature vector from the variable *features* and associates the two vectors by adding a next key-value pair into their list. Finally, it runs speech synthesis of "O.K." On the other hand, if it recognizes a question, it reads the image feature vector and going to find a corresponding index. It applies the association mechanism to calculate the vector coding a word index (by mixing the image features from keys and making an analogical mixture from values according to the collected list of the key-value pairs) and decodes it. If the index is valid, the agent runs speech synthesis of "This is a ..." and adds the word from the vocabulary at the revealed index. When the revealed index is out of range, the agent calls speech synthesis with "I have no idea." Optionally, it can do the same also when the relevance mentioned above is below the threshold. Before calling speech synthesis, the agent writes *True* into the blackboard variable *speaking* and *False* after it is done. Then the agent waits for the next text.
- *ViewerAgent* displays the image from the camera, so it regularly reads the variable *camera* and displays its content. It is good for the user as he has to learn that the shown object is within the camera's field of view. It is also helpful to display the currently recognized name. We can manage it by adding the variable *name* written by the *ControlAgent* and read by this agent. This information is valid for a limited time, so we employ the time validity mechanism supported by the blackboard. The *ControlAgent* writes the value for a limited time, after which the value automatically disappears from the blackboard. As a result, the *ViewerAgent* reads from the blackboard the default, which is an empty string in this case.
- *LipsAgent* provides movement of the robot imitating speaking. It frequently wakes up and checks the variable *speaking* value. If it is *True*, it changes the current facial expression. If it is *False*, it returns the facial expression to a neutral state, e.g., closes open mouth. As a result, the user has the impression that the robot moves its lips when hearing the synthesized speech.

We wrote almost all values of the backboard variables with reasonable time validity. Moreover, we employ the priority mechanism when necessary. For example, we can instruct the robot by saying: "This is a cup." However, we can also ask the robot: "What is this?" and the robot speaks: "This is a cup." The same sentence can have different meanings when we say it and when the robot says it. Our system is not able to recognize who is speaking. As a result, the *ControlAgent* could generate a voice that other agents record and transcribe, and it waits at the blackboard to be processed as the user says it. Traditionally we would manage this by stopping recording during calling speech synthesis. But it is not so easy since the *ListenerAgent* is mostly involved. It is much easier to inhibit data transfer between the *ListenerAgent* and the *TransciptionAgent*. So, the *ControlAgent* writes a nope sound to the variable *audio* with a higher priority when it starts the speech synthesis. After it, it writes the value again with a limited time validity. As a result, after the delay, the nope sound disappears from the blackboard, and the communication between the *ListenerAgent* and the *TranscriptionAgent* works again. In this way, we eliminate the problem easily.

| | |
|---|---|
| What is this? 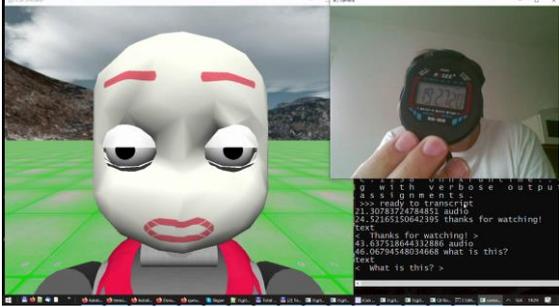 I have no idea. | This is a knife 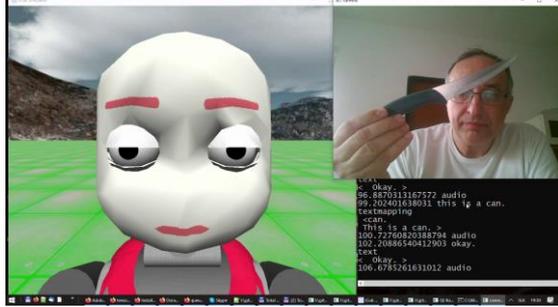 O.K. |
| This is a watch. 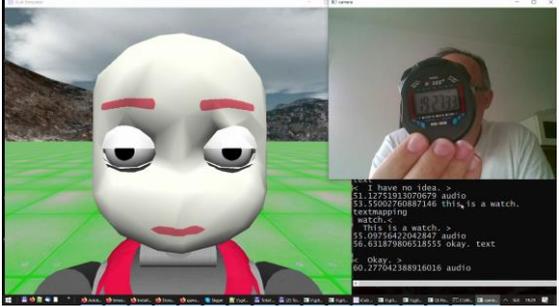 O.K. | What is this? 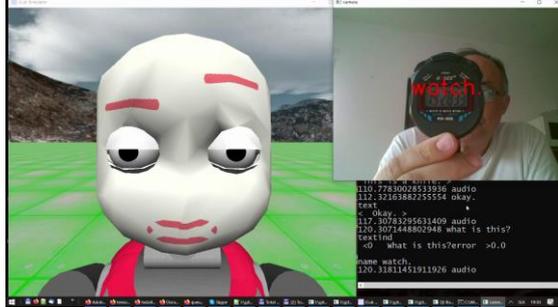 This is a watch. |
| This is a bottle. 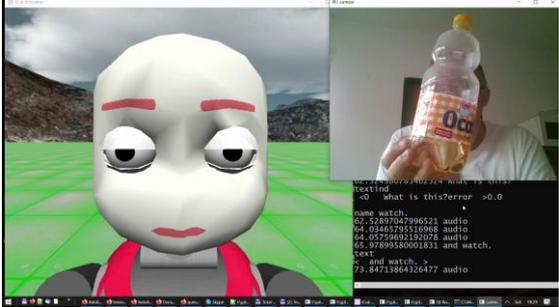 O.K. | What is this? 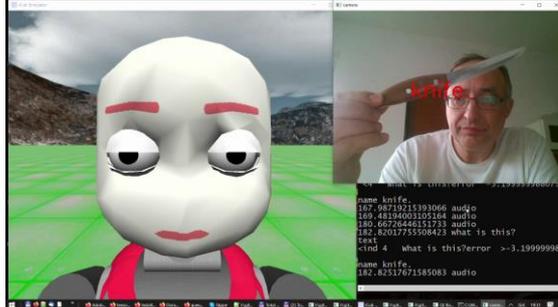 This is a knife. |
| This is a can. 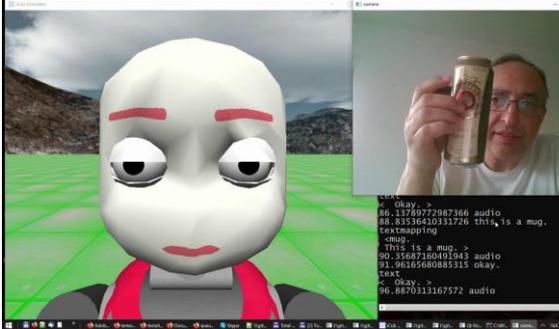 O.K. | What is this? 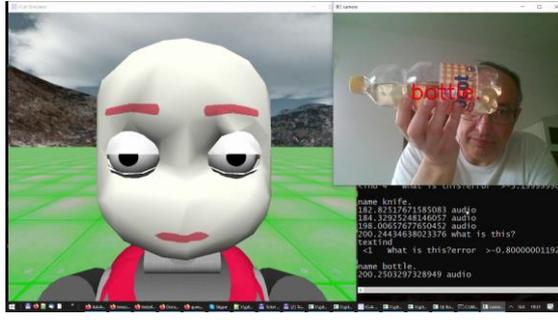 This is a bottle. |

Fig. 7. Example of one possible scenario of the system operation

## V. RESULTS

We have tested the system with a set of ordinary objects: a bottle, glass, cup, can, knife, watch, pen, bullet, medicine, and others. We had to manage two conditions: a silent room (otherwise, we fail to stop recording when the user stops speaking) and no direct light to the camera (causing poor quality of images). One possible scenario is in Figure 7. The quality is quite impressive.

The system's scalability is limited rather by the patience of the user than by the number of object categories. However, some objects have less support by the foundation model than others, e.g., scissors. Simply, our quality is as good as the quality of the model. For example, the model is trained with augmentation by rotation up to 45 degrees. As a result, sometimes it happens that a bottle named in the upright position is recognized as glass if we rotate it to the horizontal position. Of course, it is enough to call the bottle even in the horizontal position, and this misbehavior disappears.

On the other hand, if we run the mechanism offline on a dataset, we quickly lose 100% quality even when we apply the corrections. (We correct just the last pattern; if we saturate corrections, we would achieve 100%). We have prepared a dataset containing 732 pictures of 30 object categories presented on various backgrounds. Then, we name a single example for each kind of object in a randomly selected context. After each class, we evaluate all presented pictures. Optionally, we correct the name for images of the latest category. We see that corrections generally help but could negatively affect the recognition quality of former patterns (Figure 8, see green and blue curves). We see that low category numbers are safe.

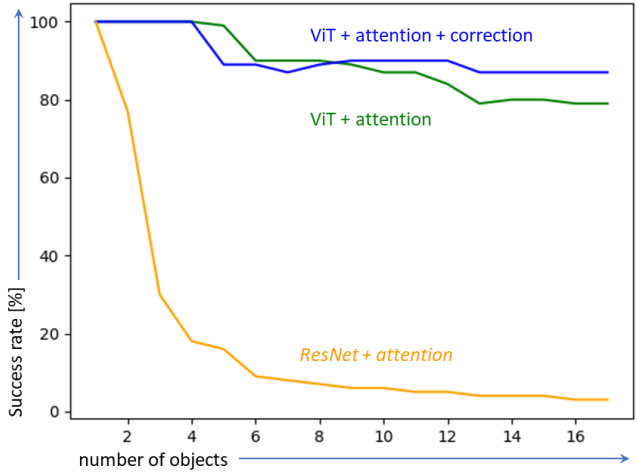

Fig. 8. Offline evaluation on a dataset

The system is pretty resistant to the change in the background (Figure 9, on the top). It is the main benefit that the user will notice. Mainly if the set of objects is wholly named under the same conditions, and just then, the background varies. We get the same positive results for calling things closer to the camera, named farther from the camera, and oppositely (Figure 9, on the bottom).

We have also tried to replace the model based on the vision transformer with the one based on the ResNet architecture (Figure 8, see green and orange curves). The two models are trained by the same method (DINO) from the same dataset. However, their suitability for our application is ultimately different. When we still have 98% with the transformer, we have 16% with ResNet. This result enlightens the advantages of the transformer architecture. On the other hand, our association mechanism could not save the ResNet backbone's worse quality, which is not convincing. It seems that the latent space provided by the transformer is much closer to linear than the ResNet's.

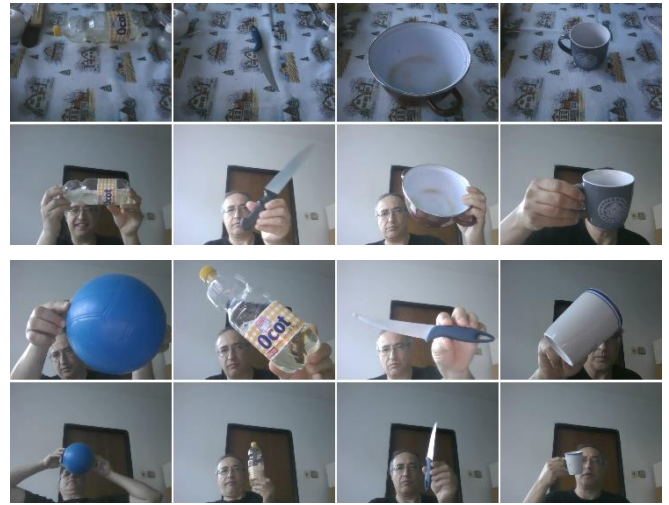

Fig. 9. Examples of varying background and distance from the camera. In both cases, we name objects in the top line, and the system succeeds in calling the ones in the bottom line and vice versa.

Teaser videos are available at https://youtu.be/nQnEdyfDh5I (stable background) and https://youtu.be/VFPgt9UyUbI (varying background).

## VI. CONCLUSION

We have implemented a control system of a robot that - in interaction with the user - learns to name objects. It provides impressive quality and works in a one-shot manner. In this way, we have demonstrated the power of the available foundation models. The system still has a potential for further improvement based on employing attention maps.

We share all codes at https://github.com/andylucny/whatisthis

### ACKNOWLEDGMENT *(Heading 5)*

This work was supported by the Horizon Europe project TERAIS, no. 101079338, and partly by the national project VEGA 1/0373/23.

## REFERENCES

[1] C. Silberer, S. Zarriess, and G. Boleda, "Object Naming in Language and Vision: A Survey and a New Dataset" Proceedings of the 12th Conference on Language Resources and Evaluation (LREC 2020), pp. 5792–5801, Marseille, May 2020

[2] A. Radford, J. W. Kim, Ch. Hallacy, A. Ramesh, G. Goh, S. Agarwal, G. Sastry, A. Askell, P. Mishkin, J. Clark, G. Krueger, and I. Sutskever, "Learning Transferable Visual Models From Natural Language Supervision", Volume 139: International Conference on Machine Learning, July 2021, Virtual,

https://doi.org/10.48550/arXiv.2103.00020

[3] T. Nishikawa, K. Aoyama, Sh. Sekiguchi, T. Takayanagi, J. Wu, Y. Ishihara, T. Kojima, and J. J. Yokono, "Naming Objects for Vision-and-Language Manipulation", March 2023, DOI: 10.48550/arXiv.2303.02871

[4] M. Zambelli, A. Cully, Y. Demiris, "Multimodal representation models for prediction and control from partial information", Robotics and Autonomous Systems 123 (2020).

https://doi.org/10.1016/j.robot.2019.103312

[5] C. Zhou, Q. Li, C. Li, J. Yu, Y. Liu, G. Wang, K. Zhang, C. Ji, Q. Yan, L. He, H. Peng, J. Li, J. Wu, Z. Liu, P. Xie, C. Xiong, J. Pei, P. S. Yu, and L. Sun, "A Comprehensive Survey on Pretrained Foundation Models: A History from BERT to ChatGPT", (2023), ArXiv. /abs/2302.09419

[6] D. Vernon, G. Metta, and G. Sandini, "The iCub cognitive architecture: Interactive development in a humanoid robot", In: IEEE 6th International Conference on Development and Learning. pp. 122–127 (2007). https://doi.org/10.1109/DEVLRN.2007.4354038

[7] M. Caron, H. Touvron, I. Misra, H. Jegou, J. Mairal, P. Bojanowski, and A. Joulin, "Emerging properties in self-supervised vision transformers". In: Proceedings of the International Conference on Computer Vision. ICCV (2021)

[8] A. Dosovitskiy, L. Beyer, A. Kolesnikov, D. Weissenborn, X. Zhai, T. Unterthiner, M. Dehghani, M. Minderer, G. Heigold, S. Gelly, et al. "An image is worth 16x16 words: Transformers for image recognition at scale." preprint arXiv:2010.11929, 2020

[9] A. Vaswani, N. Shazeer, N. Parmar, J. Uszkoreit, L. Jones, A. N. Gomez, L. Kaiser, and I. Polosukhin, "Attention is all you need." In: 31st International Conference on Neural Information Processing Systems. ACM (2017)

[10] A. Lucny, "Building complex systems with agent-space architecture", Computers and Informatics 23(1), 1–36 (2004)

[11] A. Lucny, "Towards one-shot learning via attention", In: CEUR Workshop Proceedings, ITAT 2022. pp. 4–11. 3226 (2022)

[12] Y. Yang, Y. Li, C. Fermuller, and Y. Aloimonos, "Robot Learning Manipulation Action Plans by Watching Unconstrained Videos from the World Wide Web", Proceedings of the Twenty-Ninth AAAI Conference on Artificial Intelligence, ICCV 2015

[13] A.Thomaz, and Maya Cakmak, "Interactive Learning of Object and Object-Driven Interaction from Human Teachers", RSS 2012

[14] E. Krause, M. Zillich, T. Williams, and M. Scheutz, "Learning to Recognize Novel Objects in One Shot through Human-Robot Interactions in Natural Language Dialogues", Proceedings of the AAAI Conference on Artificial Intelligence 28(1) June 2014, DOI: 10.1609/aaai.v28i1.9143

[15] M. Matarese, A. Sciutti, F. Rea, and S. Rossi, "Toward Robots' Behavioral Transparency of Temporal Difference Reinforcement Learning With a Human Teacher", IEEE TRANSACTIONS ON HUMAN-MACHINE SYSTEMS, VOL. 51, NO. 6, DECEMBER 2021

[16] J. Ambsdorf, A. Munir, Y. Wei, K. Degkwitz, H. M. Harms, S. Stannek, K. Ahrens, D. Becker, E. Strahl, T. Weber, and S. Wermter, "Effects of Explanations in Human-Robot Interaction", In: 31st IEEE International Conference on Robot and Human Interactive Communication (RO-MAN), pp. 393-400, Naples, Italy, 2022